# Detection of Critical Number of People in Interlocked Doors for Security Access Control by Exploiting a Microwave Transceiver-Array

Paolo Nesi, Gianni Pantaleo
DISIT Lab, Distributed Systems and Internet Technologies Lab.
Department of Systems and Informatics, University of Florence, Firenze, Italy
Tel: +39-055-4796523, fax: +39-055-4796363, http://www.disit.dsi.unifi.it
email: nesi@dsi.unifi.it, pantaleo@dsi.unifi.it

***Abstract*** *– Counting the number of people is something many security application focus on, when dealing with controlling accesses in restricted areas, as it occurs with banks, airports, railway stations and governmental offices. This paper presents an automated solution for detecting the presence of more than one person into interlocked doors adopted in many accesses. In most cases, interlocked doors are small areas where other pieces of information and sensors are placed in order to detect the presence of guns, explosive, etc. The general goals and the required environmental condition, allowed us to implement a detection system at lower costs and complexity, with respect to other existing techniques. The system consists of a fixed array of microwave transceiver modules, whose received signals are processed to collect information related to a sort of volume occupied in the interlocked door cabin. The proposed solution has been statistically validated by using statistical analysis. The whole solution has been also implemented to be used in a real time environment and thus validated against real experimental measures.*

## I. INTRODUCTION

The identification of human presence and people counting are fundamental tasks in security systems and, more generally, in human-computer interaction (e.g., people flow estimation and monitoring in public transport, smart interfaces etc.). The large variety of commercial sensors has involved the development of several different approaches to cope with these tasks, depending on the various features or pieces of information you are interested in, which can be extracted. The most commonly proposed solutions range from active or passive radiometry to computer vision based techniques. As to image and video information retrieval, people counting is one of the most challenging tasks in object detection, due to large variations in appearance, articulations and movements, occlusions caused by carried luggage, the presence of shadows, as well as the dependence of shape characteristics from the view point. Full automation of security systems and related tasks, such as people motion tracking under general conditions (e.g., in large spaces or outdoor environments), usually require dynamic approaches to the problem (e.g., by means of scanning sensors, mobile robots, video or image information retrieval, etc.), with better resolution, at the expense of higher production costs.

In literature, active and passive radiometry based methods have been proposed for people detection and counting, from active Doppler radar detectors [1] to passive infrared, microwave or millimeter-wave radiometers [2]. Widely used devices are Laser range finders [3], [4], which can directly measure objects' geometry and target distances in a very accurate way [5]. Three different types of algorithms can be grouped into this last mentioned category: heuristic based [6], [7], when the detection is done accordingly to a set of heuristic rules, which aim at describing significant characteristics of human body and shapes; motion based [8], [9], when the detection is performed by retrieving the differences among consecutive laser scans, or between a scan and the background environment; and finally, feature based algorithms [3], [10], which use geometric characteristics for object detection, and often work together with supervised learning algorithms, such as *AdaBoost* [11].

Furthermore, human detection, people tracking, and people counting have been often approached as feature extraction problems from videos and images. Computer vision methods are well suited for video surveillance of outdoor or large indoor spaces, and for pedestrian monitoring. In this field, the most popular solutions can be grouped into motion-based [12], [13], and shape-based techniques [14], [15]. The former generally relies on subtracting moving objects (foreground) from the background (in this case, detection performances are highly influenced by the background modeling); the latter aims at detecting people directly, according to shape information and models. Some recently proposed methods use other approaches like Viola and Jones use of Haar-like wavelet features for pedestrian detection [16], Histogram of Oriented Gradients (HOG) features [17], [18] and particle filters [19], [20], [21].

More specifically, concerning people counting tasks based on video analysis, two main approaches have been proposed: the direct method consists in first detecting people in the images (by means of some object detection techniques), and then counting them; with the indirect method, counting is performed on the basis of features which do not require a preliminary people detection [22]. Regarding the latter approach, more recently used techniques perform the measurement of moving pixels [23], blob size [24] and fractal dimension [25]. Albiol et al. in [26] suggest the use of corner points as features found using a variant of the Harris corner detector [27]. Conte et al in [22] aimed at improving this last mentioned solution, in case of groups with many people at different distances from the camera, using clustering techniques and Inverse Perspective Mapping for perspective corrections. Padua et al. in [28] have realized a method to count the pedestrian flow, relying on a combination of motion estimation, spatio-temporal segmentation and prior knowledge about body dimensions and walking speed. Del Bimbo and Nesi in [29] presented a vision-based counting people solution grounded on optical flow field estimation; the solution exploited the spatio-temporal analysis and subsequent smoothing in the feature flow domain.

Other solutions for estimating people flow are based on different types of sensors, like photoelectric [30] or pressure sensors [31], even if their use is often limited to restricted indoor spaces.

In this paper, a low cost method and solution to detect the presence of more than one person inside a secured interlocked door (e.g., for bank access), are proposed. The solution can be applied to almost all kinds of interlocked doors for access control. As described in the following section, the solution exploited sensors that are typically available in those environments and therefore, it does not increase too much the costs of the door system itself. In Section II, the main rationales and general requirements at the basis of the proposed solution are described. Then, Section III deals with an overview of the system architecture together with the rationales at the basis of the adoption of the specific sensors used in the solution. In Section IV, the algorithm for detecting critical cases where more than one person has entered the interlocked door is presented. In order to build a decision/detection rule and to validate the model, a statistical analysis has been performed. To this end, the output of a multivariate regression model (built with several collections of test data) has been chosen, which yielded better detection results than simply thresholding processed data. In Section V validation measurements and results, in different experimental conditions, are illustrated. The validation has been performed by using direct measurements made with a prototype working in real time, realized in collaboration with the MESA srl [33] (an Italian company active in security systems development). In Section VI, some implementation details adopted for the proposed algorithm are described. Finally, conclusions are drawn in Section VII.

II. REQUIREMENTS

The requirement analysis started from the analysis of the above mentioned control access scenarios. The specific problems addressed in this paper are related to the detection of the critical number of people which could be present in an interlocked door. Interlocked doors are frequently used as solutions for security access control and monitoring. They are typically located at the entrance of banks, jewelry, governmental offices, etc. Those areas and offices may be sensitive targets for both terrorism and robbery. This latter case can be considered the largest applicative market of the solution proposed. Moreover, the adoption of interlocked door is quite standard in the field of access control within restricted indoor environments. The problem refers to the particular cases of crowded environments, where people in the interlocked doors are frequently requested to identify themselves with the system by using some biometrics solutions – e.g., providing fingerprint, providing a code on a small keyboard, via voice recognition, providing a retina scan, being weighted, and/or showing a badge. A combination of these techniques is typically used to increase security level.

On the other hand, despite these multiple techniques, the authorized person may be constrained/forced to enter the interlock door by a second person, maybe under the threat of a weapon. Therefore, the attacker may hold tight the authorized person so as to enter the interlocked door, thus providing the needed information to get out of the interlocked door. Those events are not easy to be detected, since the two embraced people entering the interlocked door could be mistaken for a large size person, etc. Moreover, the identification of these critical cases has to be performed in real time (it is not possible to keep a person into the small interlocked door for several seconds). So that, once the critical case is identified, the systems has: (i) to call the security which should go to the door and control or talk with the people inside it, and/or (ii) to open the entering door and thus letting the people get out of it.

At first glance, a non-intrusive solution to detect critical cases could be based on the adoption of (i) weight sensors under the floor; but they are considered inadequate, since they could lead to false detection, when it comes to heavy-weighted people; besides, such sensors are very expensive because they need to be tuned and retuned frequently; moreover, a precise person profile is also needed, which restricts the sensor application to well-known people; (ii) vision based systems are very expensive and need to have a clean space, controlled light conditions, etc. On such grounds, an alternative was requested so as to overcome these problems and to reduce general costs.

## III. SYSTEM ARCHITECTURE OVERVIEW

On the basis of the above reported requirements and considerations, the chosen solution for the system design has been grounded to exploit a fixed array of microwave transceiver modules, to collect information related to volume variations inside the cabin. Those sensors are typically used as on/off devices for intrusion detection, while in the proposed applications they are used as tools for measuring the volume occupied under the sensors. The microwave transceiver sensors are used to build a fixed array, and are HB 100 X-band microwave modules [32]. They are mounted on the inner roof of a curved glass bank door (the above mentioned interlocked door), so that the main antenna beam is directed vertically towards the floor. The functional architecture of the model, and the geometry of the transceiver array are shown in Figure 1. The proposed solution has been grounded on exploiting the same electronic control board which acquires the sensor data for the identification of the full entrance of the person inside the interlocked door. The work presented in this paper increases the security level of interlocked doors with an additional system to detect some critical conditions.

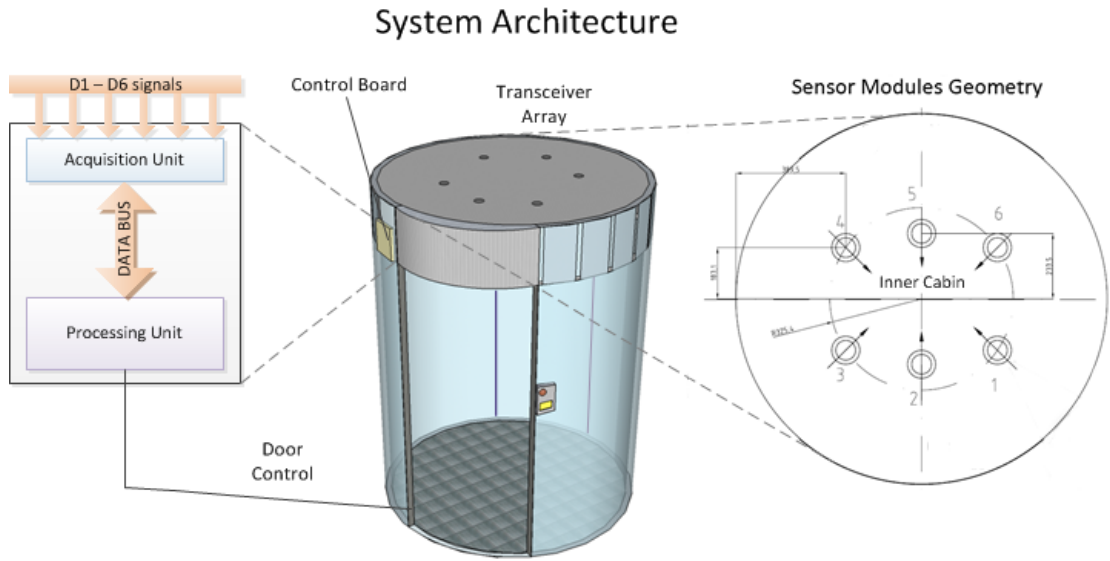

Figure 1. General view of system architecture, and detail (top view) of inner roof illustrating the sensor modules geometry. The six devices are referred to as D1,…, D6.

The working modality for those sensors is typically based on the fact that they include, in a single package, both transmission and reception drivers. Typically, these sensors are adopted to implement alarms sensors, a power signal is produced by the transmission driver and then the reception device is activated to measure the power. The performed measure is adopted to produce a binary alarm, when a given threshold is reached, thus corresponding to a presence of a movement in the room, i.e., in front of the sensor. Please note that these sensors are not used typically by measuring the power on their reception antenna with a high rate, but only taking some samples few time instances after the transmission and maybe taking the average.

When the receiving device is sampled, the trend of the received power can be obtained. In Figure 2, the typical behavior of a signal received by a single sensor in time is presented. The diagram reports a temporal window starting few seconds before the door opening (when the cabin was empty) and continue showing the passage of a person into the interlocked door. At regime, when the interlocked door is empty and entrance door is closed, the signal presents small variations around a steady-state constant value. At $t_{START}$ the door is open, thus the sensor is used to measure the received power with a high sample rate (as described in the following). The solution aims at retrieving information related to the occupied "volume", by measuring the trend of the received signal, reflected by bodies inside the cabin. The changes in power received are extracted and normalized as depicted in Figure 2 on the right side.

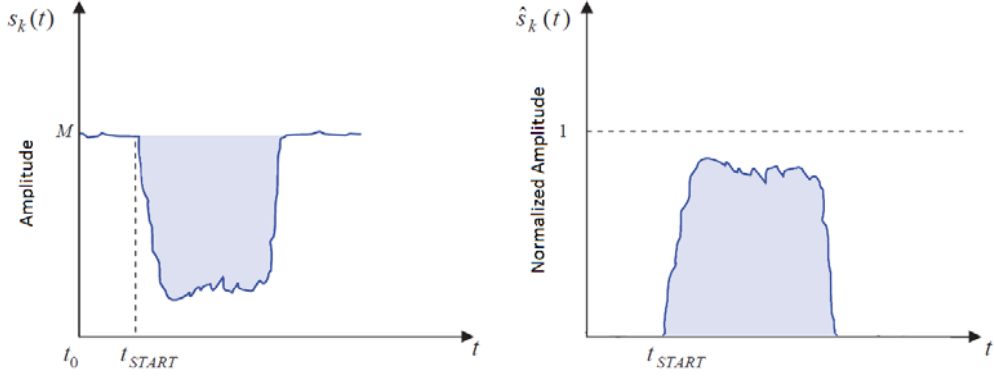

Figure 2. Example of extracted information, $\hat{s}_k(t)$, for a generic received signal $s_k(t)$; $t_{START}$ indicates the instant when the entrance door opens.

According to Figure 1, the interlocked door is equipped with 6 sensors on the roof. In our case, with the aim of detecting the presence of more than one person in the inner area, we decided to exploit all the 6 sensors to increase robustness of the measure and thus of the detection decision algorithm, as described in the next section. The information gathered through the 6 sensors is not enough to reconstruct a sort of image of the inner cabin; however, this is beyond our intents. The obtained signal could be considered strongly correlated to the volume under the sensors.

The analysis of the data provided by a single sensor (e.g., amplitude averaging and thresholding, calculation of area under the function etc.) has turned out to be not enough for the detection of critical cases, namely two people entering together. Therefore, it was necessary to identify first, and subsequently validate, some metrics directly obtained by processing jointly those signals retrieved by the six sensors (as discussed in Section IV and Section V). In addition, a statistical model was applied (based on a logistic regression analysis) to verify whether the model and the identified signals can describe the collected data in a more robust way (with a better rejection of noise, or minor sensibility to outliers). Moreover, the statistical analysis and the model validation showed the presence of a high degree of correlation of the identified measures with the expected output variable (number of people).

IV. DETECTION ALGORITHM TO IDENTIFY CRITICAL CASES

The first step for the exploitation of the sensor array has been the set up of solution for their activation and reading. To this end, the 6 sensors have been used as depicted in Table I by alternating transmissions from one device and measures of signals on the others. When a transceiver is in transmission state, the remaining five devices are in reception. The delay between two consecutive transmissions has been set to 5 ms. For each device transmitting, the signal received by the same sensor is discarded, since it is obfuscated by transmission. The acquisition board gathers these data, mapping them into a 6×6 array with empty diagonal. In the following, we refer to acquired data, received in a time frame, with the notation reported in Table I: a generic received signal is denoted with $x_{i,j}(t)$, $i = 1, ... ,6$, $j = 1, ... ,6$ and $i \neq j$.

The real-time implementation has been necessary to get the precise measures directly on the control board.

TABLE I. RECEIVED DATA MAPPING: NUMERICAL ID AND SIGNAL NOTATION. SIGNALS UPON THE DIAGONAL ARE NOT PROCESSED.

| Mapping of signals acquired from the six transceivers array | | | | | | |
|---|---|---|---|---|---|---|
| Rx device / Tx device | D1 | D2 | D3 | D4 | D5 | D6 |
| D1 | - | id = 1 $x_{1,2}$ | id = 2 $x_{1,3}$ | id = 3 $x_{1,4}$ | id = 4 $x_{1,5}$ | id = 5 $x_{1,6}$ |
| D2 | id = 6 $x_{2,1}$ | - | id = 8 $x_{2,3}$ | id = 9 $x_{2,4}$ | id = 10 $x_{2,5}$ | id = 11 $x_{2,6}$ |
| D3 | id = 12 $x_{3,1}$ | id = 13 $x_{3,2}$ | - | id = 15 $x_{3,4}$ | id = 16 $x_{3,5}$ | id = 17 $x_{3,6}$ |
| D4 | id = 18 $x_{4,1}$ | id = 19 $x_{4,2}$ | id = 20 $x_{4,3}$ | - | id = 22 $x_{4,5}$ | id = 23 $x_{4,6}$ |
| D5 | id = 24 $x_{5,1}$ | id = 25 $x_{5,2}$ | id = 26 $x_{5,3}$ | id = 27 $x_{5,4}$ | - | id = 29 $x_{5,6}$ |
| D6 | id = 30 $x_{6,1}$ | id = 31 $x_{6,2}$ | id = 32 $x_{6,3}$ | id = 33 $x_{6,4}$ | id = 34 $x_{6,5}$ | - |

As an example, $x_{1,2}$ refers to the signal received by D2 device when the only device D1 is transmitting. A 30 ms time resolution frame is chosen, which is sufficient to collect received data for a single transmission. This is equivalent to consider that each row of the array map in Table I is acquired, that there is a 5 ms gap between the acquisition of two consecutive rows, and that the whole array is acquired in 30 ms.

The main phases of the detection algorithm are:

- *Pre-Processing* to perform basic reading operations, normalization and summation of useful portions extracted from received signals.
- *Calibration* to compute the calibration factors, which are used for normalization of signal and for the estimation of metrics, respectively.
- *Metrics estimation* to perform the estimation of a set of measures that could be used for decision taking mechanisms in the successive phase according to the developed model. Among the measures the information about the pseudo volume occupied under the sensors, the possible variance of the pseudo volume information that should be an index of the fragmentation of the objects in the interlocked door.

- *Decision* to discern critical cases where there might be more than one person in the interlocked door. The detection algorithm is based on the parameters estimated by the statistical analysis as described in the next Section V.

*A. Pre-Processing*

In this phase, preliminary operations are performed on the sensor data. The reading buffer receives, as input, the $s_{in}(t)$ data flow, structured as follows:

$$s_{in}(t) = [b_s,\ n_{ID}^x,\ x_{i,j}(t)];$$

where $x_{i,j}(t)$ and $n_{ID}^x$ are the received signals and their corresponding numerical ids (according to Table I); $b_s$ is a status describing the condition of the inner cabin (presence or absence of people in the inner cabin, entrance door closed or open, exit door closed or open). In this phase of preprocessing, it is also possible to identify the temporal windows and thus the conditions in which the Calibration can be performed. These conditions correspond to door closed and nobody is inside, see for example Figure 2, before any starting.

Besides, once calibration factors have been computed or updated, the *Pre-Processing* carries out signals normalization and subsequent summation. Finally, data resulting from this last operation are sent to the estimation of *Metrics* and thus to the *Detection* phase.

*B. Calibration*

As long as the interlocked door is empty (nobody is inside the cabin), the signals received by D1 – D6 devices appear like the ones in the example below, as depicted in Figure 3. Figure 3 presents five time plots of signals received by D1, D2, D3, D4 and D6 devices when D5 is transmitting; the cabin is empty and both entrance and exit doors are closed. Under these conditions, the signals assume values with very small amplitude variations. These values are used for normalization of each acquired data array, and for calculation of decision model (see section IV.C).

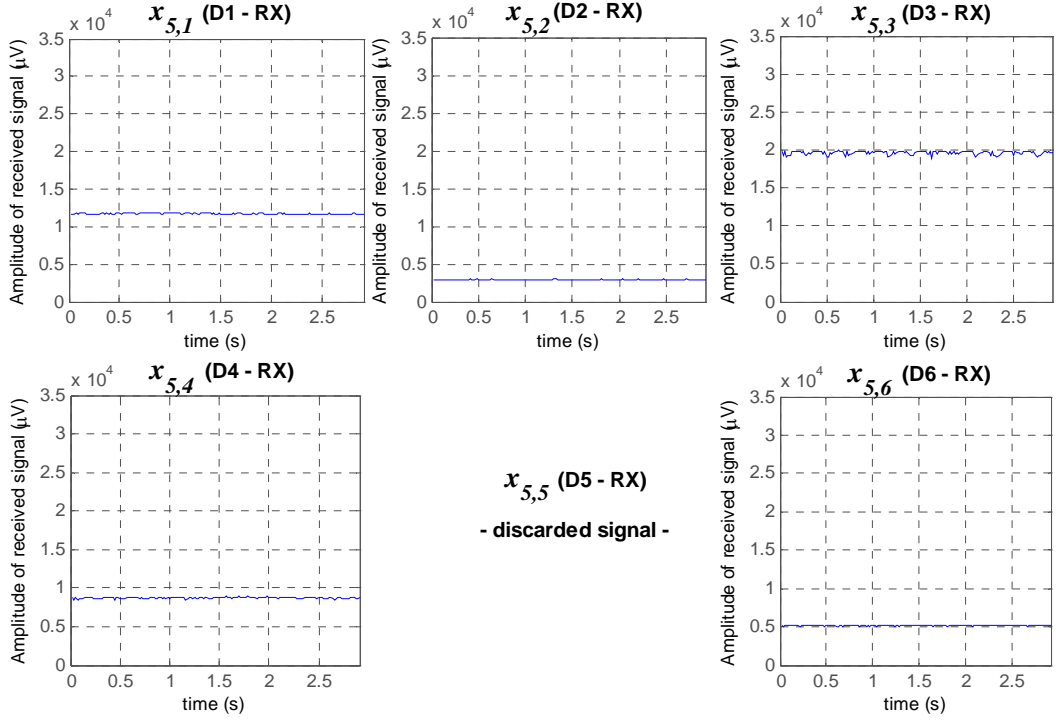

Figure 3. The time plot of received signal arrays by D1 – D6 devices, while D5 is in transmission state, thus the graph is missing for the reception. The inner cabin of the interlocked door is empty, there are no people inside the scanned area. Signal notation refers to Table I.

The calibration factors (one for each combination: transmission reception) are defined as:

$$\hat{c}_{i,j} = E\{x_{i,j}^E(1,\dots,N_E)\}, \tag{1}$$

where $x_{i,j}^E(t)$ are the received signals when the cabin is empty; $N_E$ is the time index, which controls the number of signal samples used for the estimation of the calibration factor of Eq. (1). Please see Section VI for more implementation details.

*C. Metrics Estimation*

The significant measures can be taken into the temporal windows where booth doors are closed. The control system allows to bring those signals into the control board. The measuring system receives a signal when the door is open. According to Figure 4 and Figure 5, it is evident that the signal regarding the pseudo volume measured by a sensor presents several fluctuations which are due to the movements and change of geometry of the objects under the sensors.

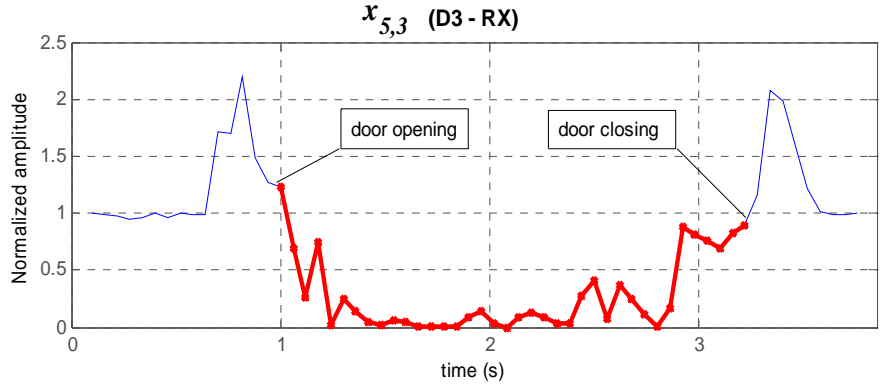

Figure 4. An example of the useful signal portions used in the next stages processing is highlighted in red (please note that this signal has been taken before calibration as in Figure 2 left).

Thus, the signal portions of interest, $x_{i,j}^F(t)$, are represented by those values taken in the frames between the closure of entrance door and the opening of the exit door (see Figure 5). These extracted signals are normalized by using the calibration factors (1), thus obtaining:

$$x_{i,j}^N(t) = 1 - \frac{x_{i,j}^F(t) - min\{x_{i,j}^F(t)\}}{\hat{c}_{i,j}}, \quad (2)$$

where $x_{i,j}^N(t)$ denotes the normalized and reversed portions of the received signals.

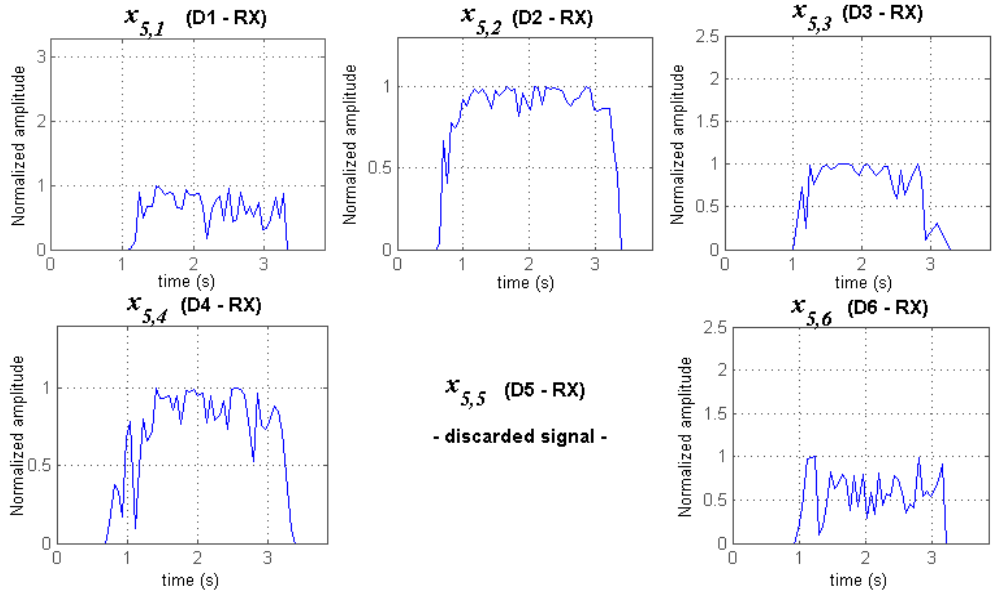

Figure 5. Time plot of received (and pre-processed) signal arrays by D1 – D6 devices, while D5 is in transmission state. There are two people in the inner cabin. The pre-processing consists in extracting portions of interest as to the received signals.

Relying on the sensors behavior and on experimental observations, the following four metrics have been defined:

- **M1**: pseudo volume information, computed by summing the contributes of signals received by all the sensors, when a single device is in transmission. It is defined as $\hat{V}_{TX}$ in the following;
- **M2**: statistical variance calculated on the pseudo volume data providing $\hat{V}_{TX}$. It is defined as $\hat{\sigma}^2_{V_{TX}}$ in the following;
- **M3**: pseudo volume information, computed by summing the contributes of signals received by a single sensor, when the other devices are in transmission. It is defined as $\hat{V}_{RX}$ in the following;
- **M4**: statistical variance calculated on the pseudo volume data providing $\hat{V}_{RX}$. It is defined as $\hat{\sigma}^2_{V_{RX}}$ in the following.

*Metrics M1 and M2:*

Processed signals are summed across the rows of mapping matrix in Table I, which means summing all the signals received for a single antenna transmission (whether referring to metrics description, this case leads to the definition of M1 and M2):

$$S_i^{TX}(t) = \sum_{j=1}^{6} x_{i,j}^N(t), \quad i = 1, \dots, 6. \tag{3}$$

Thus, all the $S_i^{TX}(t)$ values are further summed together, in order to gather information related to the inner pseudo volume occupied:

$$V_{TX}(t) = \sum_{i=1}^{6} S_i^{TX}(t). \tag{4}$$

The area of the time-varying pseudo volume profile $V_{TX}(t)$ is computed again by summation:

$$\hat{V}_{TX} = \sum_{t=1}^{N_F} V_{TX}(t), \tag{5}$$

where $N_F$ is the frame length of the processed signals ($x_{i,j}^N$, $S_i$ and $\hat{V}_{TX}$).

The *Variance* $\hat{\sigma}^2_{V_{TX}}$: is estimated as

$$\sigma^2_{V_{TX}} = Var(\hat{V}_{TX}) \tag{6}$$

*Metrics M3 and M4:*

Similarly to M1 and M2, signals may be summed across the columns of the mapping matrix of Table I, in order to obtain volume related information about a single receiver, when the other devices are in transmission mode.

$$S_j^{RX}(t) = \sum_{i=1}^{6} x_{i,j}^N(t), \quad j = 1, \ldots, 6;$$

$$V_{RX}(t) = \sum_{j=1}^{6} S_j^{RX}(t);$$

$$\hat{V}_{RX} = \sum_{t=1}^{N_F} V_{RX}(t). \tag{7}$$

$$\sigma_{V_{RX}}^2 = Var(\hat{V}_{RX}). \tag{8}$$

In this latter case, the following assumption is made: the time interval of 5 ms delay between consecutive transmissions is considered not relevant against possible human body movements, which may cause significant variations on received signals' profile.

The four identified and reported metrics: $\hat{V}_{TX}$, $\hat{V}_{RX}$, $\hat{\sigma}_{V_{TX}}^2$ and $\hat{\sigma}_{V_{RX}}^2$ have been defined and computed for a high number of testing cases, monitoring the passages in the doors. The analysis of the experimental results has highlighted strong difficulties in decision taking only on the basis of one of those metrics, although a certain discriminability between the case of one and more than one person was already self-evident. For example, the trend of $\hat{V}_{TX}$, is illustrated in the example in Figure 6, where a given number of passing people into the interlocked door is reported, covering correct and critical cases.

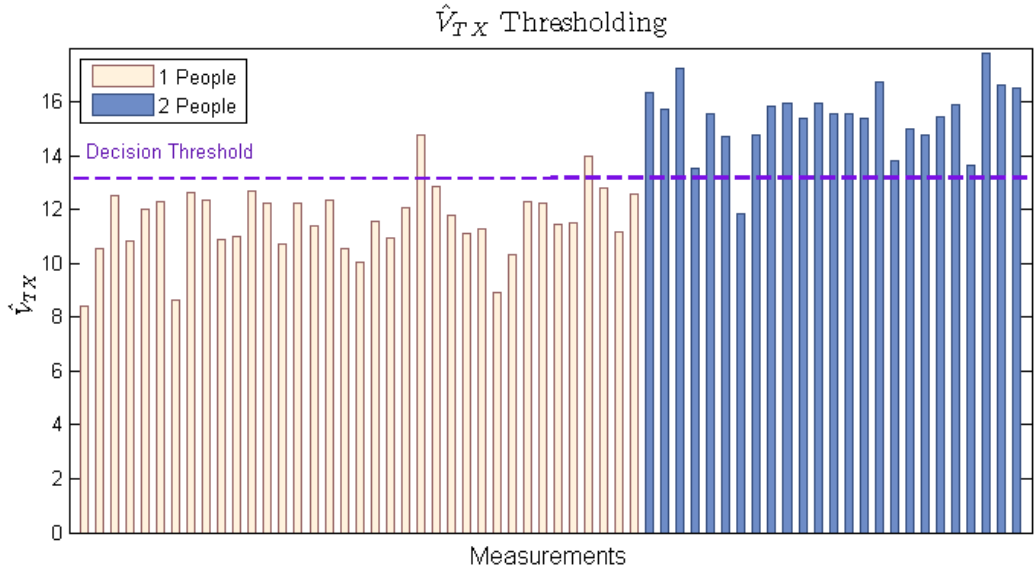

Figure 6. Threshold for $\hat{V}_{TX}$ over a mixed set of test measurement (one and two people) accessing the inner cabin.

For this reason, a statistical analysis has been performed in order to verify if a combination of the above mentioned metrics can be adopted to define a decision model. The statistical analysis and the regression models here used are reported in the next section together with the model validation. The validation of the regression model allows to identify which are the significant metrics for the identification of the critical cases and the coefficients of the identified model.

## V. STATISTICAL ANALYSIS OF THE EXPERIMENTAL RESULTS AND VALIDATION

In order to assess the possibility of using the identified metrics to create a decision model, a large data set has been produced. For each passage of people into the doors a description of those people has been recorded as well. Particular attention has been paid to include tricky cases in one person measurements (e.g., heavy build person, single person with different types of bags and baggage, overcoats etc…), and two people measurements (e.g., two tall / thin persons drawn close together). In previous Section, the problematic task of defining a decision threshold, on the values of $\hat{V}_{TX}$ and $\hat{V}_{RX}$ has been mentioned. In both cases, illustrated in Figure 6, the best threshold value which can be applied, to discriminate the number of people in the inner cabin, leads to unsatisfactory detection results.

Regression analysis is a useful technique, which aims at modeling the trend of an expected variable *Y* by means of some explicative variables $X_1, X_2, \ldots, X_N$. This is equivalent to search for appropriate coefficients $\beta_0, \beta_1, \ldots, \beta_N$ so that the output of the designed model, $\hat{Y}$, best approximates *Y* (usually minimizing the mean-squared error between *Y* and $\hat{Y}$) by a linear model, of the following form:

$$\hat{Y} = \beta_0 + \beta_1 X_1 + \beta_2 X_2 + \cdots + \beta_N X_N.$$

In our case, $\hat{V}_{TX}$, $\hat{\sigma}^2_{V_{TX}}$ and $\hat{V}_{RX}$ resulted to be significant variables in Fisher's *F*-Test and Student's *T*-Test. For further information about regression analysis theory, please refer to [34] and [35]. As a conclusive step, if $\hat{Y}$ results to be greater than an experimentally defined *Thres* value, the system detects the presence of more than one person in the inner cabin, and the output of the system is at low level, so that the door remains closed. Otherwise, the presence of one person is estimated, the system output is at high level, and the door is opened. Different types of regression models have been applied, with the purpose of identifying the most suitable one which provides the best detection results. Besides simple linear regression, robust regression analysis has been accomplished based on iterative application of LMSE method between model and expected values. By this process, the model achieves more stability against outliers, but still not satisfactory. The liner regression obtained a correlation of the about 81.58%, while for robust regression 82.33%.

*A. Validation of the Logistic Regression Model*

In this case, the expected variable may be considered as dichotomous or binary i.e., it can assume only two mutually exclusive values (i.e., *true* and *false*), thus it is suitable for logistic regression [36]. Our expected variable *Y* can assume only two independent values, depending on the presence of one person or more people within the doors. In the following, we make the assumption that $Y = 0$ when one person is present in the cabin, and $Y = 1$ when there are more than one person; by applying this model, we are interested in the probability that the expected variable takes one of these two values. In order to restrict the values range (from $[-\infty, +\infty]$ of linear regression model to $[0, 1]$ of logistic model), the expected variable can be transformed by using the *logit* function:

$$logit(P) = \ln\left(\frac{P}{1-P}\right) = \beta_0 + \beta_1 X_1 + \beta_2 X_2 + \cdots + \beta_N X_N, \quad (9)$$

where $P = \Pr(Y = 1|\underline{X})$ expresses the probability for the $Y$ variable to occur. The middle term within the logarithmic function of Eq. (9) i.e., the ratio of $\Pr(Y = 1|\underline{X})$ and $\Pr(Y = 0|\underline{X})$, is defined also as $odds(P)$.

The statistical analysis for the validation of our solution has been performed on the basis of some hundreds of test cases. From the analysis, it turned out that the logistic regression produced accuracy improvements for the discrimination of possible critical cases. For the validation of the logistic regression analysis, significance tests are performed by evaluating different parameters, with respect to linear regression, both for the good suitableness of the general model and for single variables, as well. In our case, we used the Hosmer-Lemeshow [37] test and the Wald test, [38], [39], respectively as reported in Table II and Table III. When comparing these two tables, we see that only $\hat{V}_{TX}$ and $\hat{V}_{RX}$ turn out to be significant, while $\hat{\sigma}^2_{V_{TX}}$ and $\hat{\sigma}^2_{V_{RX}}$ can be identified as not significant (actually, the p-values related to $\hat{V}_{TX}$ and $\hat{V}_{RX}$ are about 1%, while for $\hat{\sigma}^2_{V_{TX}}$ and $\hat{\sigma}^2_{V_{RX}}$ they are larger than 30%); therefore, they have been discarded as explicative variables of the model. Subsequently, the general goodness of fit for the logistic regression model improves whenever significant variables only are included (as demonstrated by the increasing Chi-Square distribution value for the Hosmer-Lemeshow test).

TABLE II. GENERAL MODEL AND SINGLE EXPLICATIVE VARIABLE SIGNIFICANCE TEST (ALL THE FOUR EXPLICATIVE VARIABLES ARE INCLUDED)

| *General Model Goodness of Fit* | | |
|---|---|---|
| **Hosmer Lemeshow Test** | Chi-Square value | p-value |
| | 10.021 | 0.264 |
| *Single Variables Significance Test* | | |
| **Decision Variables** | **Wald Test** | **p-value** |
| $\hat{V}_{TX}$ | 6.403 | 0,011 |
| $\hat{V}_{RX}$ | 6.486 | 0.011 |
| $\hat{\sigma}^2_{V_{TX}}$ | 1.024 | 0.312 |
| $\hat{\sigma}^2_{V_{RX}}$ | 0.971 | 0.325 |

TABLE III. GENERAL MODEL AND SINGLE EXPLICATIVE VARIABLE SIGNIFICANCE
TEST (ONLY THE TWO SIGNIFICANT VARIABLES ARE INCLUDED)

| General Model Goodness of Fit | | |
|---|---|---|
| Hosmer Lemeshow Test | Chi-Square value | p-value |
| | 13.857 | 0.086 |
| Single Variables Significance Test | | |
| Decision Variables | Wald Test | p-value |
| $\hat{v}_{TX}$ | 5.603 | 0.018 |
| $\hat{v}_{RX}$ | 5.693 | 0.017 |

*B. System Classification Performance*

For the general assessment of results, traditional metric models have been defined to measure classification performances of a system as described in [40]. Let us recall our dichotomous expected variable *Y*, which can assume two distinct values, depending on whether there are one or more people within the doors. It is now convenient to distinguish the following cases:

- **A) Actual Status**:
    - Case 1A: The system receives one person within the doors;
    - Case 2A: The system receives more than one person, the critical case to be detected.
- **B ) System Detection**:
    - Case 1B: The presence of one person within the door is detected;
    - Case 2B: The presence of more than one person is detected, the critical case to be detected.

The case of less than one person (namely, empty cabin) has not been considered; in fact, the use of steady-state sensor data when nobody is inside, the cabin has already been discussed in previous sections. In Table IV and Table V, the classification of validation measures, system detection performances and fault-proneness tendencies are illustrated. In Table IV, the classification has been applied to a testing set of 63 passages of one or more people according to the combination of the above mentioned occurrences.

TABLE IV. CLASSIFICATION OF VALIDATION MEASURES.

|  |  | System Detection | | |
|---|---|---|---|---|
|  |  | Case 1B | Case 2B | TOT. |
| Actual Status | Case 1A | (a) 35 | (b) 2 | (a + b) 37 |
| Actual Status | Case 2A | (c) 0 | (d) 26 | (c + d) 26 |
|  | TOT. | (a + c) 35 | (b + d) 28 | (N) 63 |

The resulting outcomes can be grouped into the following known categories of matching/error classification:

- **True Positives (TP)**: in our case, this parameter represents the number of events verifying both cases 1A and 1B, i.e., the presence of one person, correctly detected by the system (referred as $(a)$ in Table IV).
- **True Negatives (TN)**: it represents the number of events verifying both cases 2A and 2B, i.e., the presence of more than one person, correctly detected by the system (referred as $(d)$ in Table IV).
- **False Positives (FP)**: it represents the number of events for which the actual presence of one person is erroneously mistaken as the presence of more than one person (referred as $(b)$ in Table IV).
- **False Negatives (FN)**: it represents the number of events for which the actual presence of more than one person is erroneously mistaken as the presence of one person (referred as $(c)$ in Table IV).

TABLE V. SYSTEM DETECTION PERFORMANCES AND FAULT-PRONENESS TENDENCIES.

|  |  | Parameters | Percentages (%) |
|---|---|---|---|
| # *One person* correctly detected (**TP**): | 35 | Sensibility: $a/(a+c)$ | 100% |
| # *Two people* correctly detected (**TN**): | 28 | Correct Classification: $(a+d)/N$ | 96.8% |
| # Measurements with *one person*: | 37 | Correctness: $d/(b+d)$ | 92.9% |
| # Measurements with *two people*: | 26 | Completeness: $d/(c+d)$ | 100% |
| # One person mistaken as two people (**FP**): | 2 | False Positives: $b/N$ | 3.2% |
| # Two people mistaken as one person (**FN**): | 0 | False Negatives: $c/N$ | 0% |

In conclusion, we intend to demonstrate improvements, achieved with the application of logistic regression analysis, by comparing the correlation values that explicative variables ($\hat{V}_{TX}$, $\hat{V}_{RX}$ and $\hat{\sigma}^2_{V_{RX}}$) have with respect to the expected variable, $Y$, and to the output of regression model, $\hat{Y}$. These results are shown in Table VI. Comparison of correlation between expected values and the output of the different regression models, which justifies the choice of logistic regression among other possible models, is shown in Table VII.

TABLE VI. CORRELATION AMONG SIGNIFICANT EXPLICATIVE VARIABLES, EXPECTED VALUES AND THE OUTPUT OF LOGISTIC REGRESSION MODEL.

| Correlation Values | $\ell_{TX}$ | $\ell_{RX}$ | Log. Reg. Output |
|---|---|---|---|
| $\ell_{TX}$ | 1 | | |
| $\ell_{RX}$ | 0.9597 | 1 | |
| Regression Output | 0.9518 | 0.9676 | 1 |
| Expected Values | **0.8127** | **0.7911** | **0.8335** |

TABLE VII. COMPARISON OF CORRELATION VALUES (EXPRESSED IN PERCENTAGE) BETWEEN EXPECTED VALUES AND DIFFERENT REGRESSION MODELS.

| Correlation Values | Linear Regression | Robust Regression | **Logistic Regression** |
|---|---|---|---|
| Expected Values | 81.58 % | 82.33 % | **83.35 %** |

## VI. IMPLEMENTATIVE DETAILS

The general functional architecture of the system is illustrated in Figure 7. The data received by the transceivers are carried to the reading buffer by a data bus, together with some additional pieces of information: a status byte and numerical ids associated to the signals. There are four functional blocks:

- The *Pre-Processing Module* performs basic reading operations, normalization and summation of useful portions extracted out of the received signals.

- The *Calibration Module* computes the calibration factors, which are used for the signal normalization and the detection decision rule, respectively.

- The *Volume Summation* estimates the identified metrics; on the other hand, the variance estimation has been turned off in the final system solution.

- The *Decision Module* finally outputs the decision for discriminating the presence of one or more people, by performing a linear combination, and subsequent decision variables chosen after applying a multivariate regression analysis over the signal summation data. The variance of time-varying signal profiles is also calculated.

- The *Control Module* is in charge of calling all the functions built in the source code modules. It also controls a reset function which deallocates historical variables and the reading buffer. This module is composed only by simple control functions and parameters assignation; therefore, there is no real need to describe it with further details in the next subsections, as it is required, instead, for the other modules.

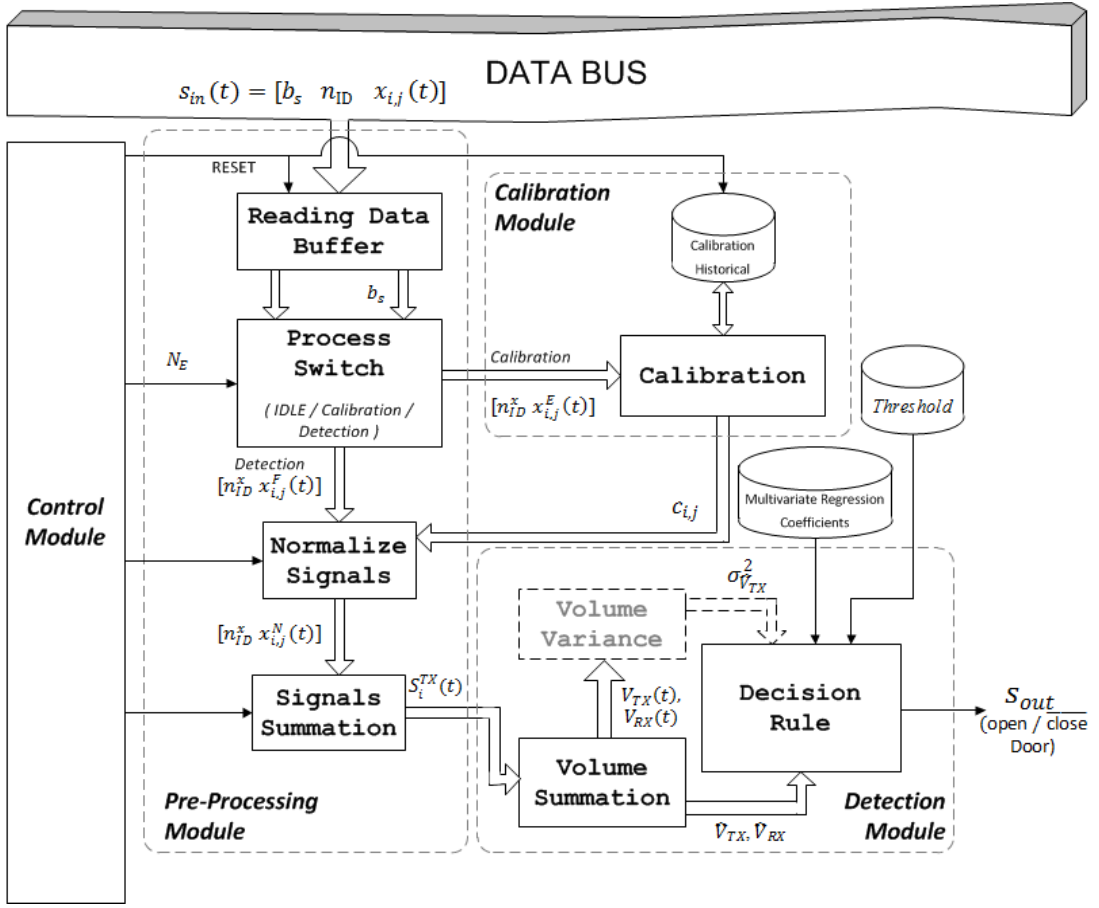

Figure 7. Block architecture of system real-time implementation.

The source code has been firstly implemented in Matlab environment, and subsequently traduced into Ansi C, to be directly programmed on the electronic board. These requirements, in addition to the limited hardware space of 24 KB on the processing board, requested static memory allocation, passing variables by address, real-time processing and iterative calculation strategies, in order to minimize variable copying and memory allocation.

An iterative algorithm has been chosen to compute the statistical mean, which is suitable for real-time implementation and minimization of allocated memory requirements. Therefore, calibration factors, defined in Eq. (1), are computed using an iterative mean algorithm, taking the following form:

$$c_{i,j}(t) = \frac{t-1}{t} c_{i,j}(t-1) + \frac{1}{t} x_{i,j}^E(t), \quad t = 2, \dots, N_E; \tag{10}$$

$$c_{i,j}(1) = 0.$$

Comparing Eq. (1) and Eq. (10), we can affirm that $\hat{c}_{i,j} = c_{i,j}(N_E)$. In a similar way, also the variance metrics $\hat{\sigma}^2_{V_{TX}}$ and $\hat{\sigma}^2_{V_{RX}}$, as defined in Eq. (6) and Eq. (8), is calculated by means of an iterative procedure, given by:

$$\sigma^2_{V_{TX}}(t) = \frac{t-1}{t}\sigma^2_{V_{TX}}(t-1) + \frac{1}{t-1}\left(V_{TX}(t) - \mu_{V_{TX}}(t)\right)^2, \tag{11}$$

$$\sigma^2_{V_{RX}}(t) = \frac{t-1}{t}\sigma^2_{V_{RX}}(t-1) + \frac{1}{t-1}\left(V_{RX}(t) - \mu_{V_{RX}}(t)\right)^2, \tag{12}$$

$$t = 2, \dots, N_F.$$

$\mu_{V_x}(t)$ and $\mu_{V_{RX}}(t)$ are the statistical mean of $V_{TX}$ and $V_{RX}$, respectively, estimated using the same algorithm used for calibration factors in Eq. (10):

$$\mu_{V_{TX}}(t) = \frac{t-1}{t}\mu_{V_{TX}}(t-1) + \frac{1}{t}V_{TX}(t), \quad t = 2, \dots, N_F; \tag{13}$$

$$\mu_{V_{RX}}(t) = \frac{t-1}{t}\mu_{V_{RX}}(t-1) + \frac{1}{t}V_{RX}(t), \quad t = 2, \dots, N_F; \tag{14}$$

We have that $\hat{\sigma}^2_{V_{TX}} = \sigma^2_{V_{TX}}(N_F)$ and $\hat{\sigma}^2_{V_{RX}} = \sigma^2_{V_{RX}}(N_F)$.

## VII. Conclusions

In this paper, a method for detecting the presence of more people in an interlocked door access has been presented. The system architecture was designed with the goal of minimizing the implementation costs. For this reason, a fixed microwave transceiver array has been used, in order to collect information related to the volume occupied in the inner cabin. Results arising from experimental tests have demonstrated that microwave transceiver modules can be profitably used as sensors to detect any possible critical conditions. As to the decision/detection rule, a simple procedure upon explicative variables has been improved by applying a logistic regression model.

Future work guidelines could include a smarter management of the detection model according to possible changes of calibration factors, depending on environmental variations of steady-state sensor values. Finally, the behavior of the system in the case of more than one person detection can be refined: for instance, while the door is kept closed, the whole estimation process could be repeated, and only if the second measure confirms the presence of more than one person, an alarm signal can be generated.

VIII. ACKNOLEDGEMENTS

The authors would like to say thanks to MESA srl and to the people who supported this research activity, providing test beds and the general infrastructure. A special 'thank you' goes to the staff involved in the test cases' set up and monitoring activities to collect data for validation.